\documentclass[letterpaper, 10 pt, journal, twoside]{ieeetran}

\usepackage[OT1]{fontenc}
\usepackage{mathrsfs}
\usepackage{stmaryrd}
\usepackage{float, cite}
\usepackage{graphicx,subfigure}
\graphicspath{{chapter/}{figures/}}
\usepackage{wrapfig} 
\usepackage{changepage}
\usepackage{color}
\usepackage{amsmath,amssymb}
\usepackage{bm,bbm}
\usepackage{multicol}
\usepackage{enumerate}
\usepackage[linesnumbered,boxed,ruled,commentsnumbered]{algorithm2e}
\makeatother

\IEEEoverridecommandlockouts
\newtheorem{problem}{Problem}
\newtheorem{assumption}{Assumption}
\newtheorem{theorem}{Theorem}
\newtheorem{corollary}{Corollary}

\newtheorem{definition}{Definition}
\newtheorem{remark}{Remark}
\newtheorem{example}{Example}

\newcommand{\sgn}{\text{sgn}}

\begin{document}

\markboth{IEEE Robotics and Automation Letters. Preprint Version. Accepted NOVEMBER, 2023}
{L. Tan \MakeLowercase{\textit{et al.}}: Motion Control of Two Mobile Robots under Allowable Collisions}

\author{Li Tan$^{1}$, Wei Ren$^{2}$, Xi-Ming Sun$^{2}$, Junlin Xiong$^{1}$
	\thanks{Manuscript received: June, 22, 2023; Revised October, 3, 2023; Accepted November, 4, 2023.}
	\thanks{This paper was recommended for publication by Editor Jaydev P. Desai upon evaluation of the Associate Editor and Reviewers' comments.
	This work was supported by the Fundamental Research Funds for the Central Universities under Grant DUT22RT(3)090, and the National Natural Science Foundation of China under Grants 61890920, 61890921 and 62273320.}
	\thanks{$^{1}$L. Tan and J. Xiong are with Department of Automation, University of Science and Technology of China, Hefei 230026, China. {\tt\footnotesize tlllll9@mail.ustc.edu.cn, xiong77@ustc.edu.cn}}
	\thanks{$^{2}$W. Ren and X.-M. Sun are with Key Laboratory of Intelligent Control and Optimization for Industrial Equipment of Ministry of Education, Dalian University of Technology, Dalian 116024, China. {\tt\footnotesize wei.ren@dlut.edu.cn, sunxm@dlut.edu.cn}}
	\thanks{Digital Object Identifier (DOI): see top of this page.}}

\title{Motion Control of Two Mobile Robots under Allowable Collisions}
	
\maketitle

\begin{abstract}
    This letter investigates the motion control problem of two mobile robots under allowable collisions. Here, the allowable collisions mean that the collisions do not damage the mobile robots. The occurrence of the collisions is discussed and the effects of the collisions on the mobile robots are analyzed to develop a hybrid model of each mobile robot under allowable collisions. Based on the effects of the collisions, we show the necessity of redesigning the motion control strategy for mobile robots. Furthermore, impulsive control techniques are applied to redesign the motion control strategy to guarantee the task accomplishment for each mobile robot. Finally, an example is used to illustrate the redesigned motion control strategy.
\end{abstract}

%%%%%%%%%%%%%%%%%%%%%%%%%%%%%%%%%%%%%%%%%%%%%%%%%%%%%%%%%%%%%%%%%%%%%%%%%%%%%%%%%%%%%%%%%%%%%%%%%%%%%%%%%
\section{Introduction}
\label{sec-intro}
%%%%%%%%%%%%%%%%%%%%%%%%%%%%%%%%%%%%%%%%%%%%%%%%%%%%%%%%%%%%%%%%%%%%%%%%%%%%%%%%%%%%%%%%%%%%%%%%%%%%%%%%%

\IEEEPARstart{M}{obile} robots have attracted increasing attention due to numerous applications in many fields including manufacturing, transportation and medicine \cite{Kagan2019autonomous}. For mobile robots, a fundamental problem is the motion control design, which has been and is still a main subject of numerous research studies. In the motion control, an essential goal is to find a controller to enable the mobile robot to achieve the target task while avoiding collisions. That is, both the collision avoidance and the target task need to be achieved simultaneously, and the collision avoidance is placed in a higher priority. 

Many approaches have been proposed to solve the motion control problem; see \cite{Kagan2019autonomous} and references therein. The energy function based approach describes obstacles by developing different energy-like functions, which are further combined with Lyapunov functions to design feedback controllers. The energy-like functions include control barrier functions \cite{RobustCBF}, artificial potential functions \cite{ImprovedAPF} and navigation functions \cite{Feedback1, Feedback2}. The optimization-based approach ensures the collision avoidance by using position constraints, and designs optimal controllers by minimizing the cost in terms of time \cite{distance1} and resource \cite{distance2,CC}. The position constraints include distance constraints \cite{distance1,distance2}, 
chance constraints \cite{CC} and conditional value-at-risk constraints \cite{CVaR}. Although the above approaches have been proposed and applied to avoid collisions, it is hard to claim that the collision avoidance can be always ensured. 
In \cite{CVaR,CBFforU,dec-iSCP,ModelingError}, the mobile robot may still collide with obstacles or other mobile robots under the controller designed from the above approach. This phenomenon results from the limitation of existing control methods \cite{CVaR,CBFforU}, the increase in the number of mobile robots \cite{dec-iSCP} and systematic modeling error \cite{ModelingError}. However, the occurrence of collisions is not necessarily bad for the task accomplishment. If the forces from the collisions can be accepted by the mobile robots \cite[Chapter 6]{CollegePhysics}, then the collisions have no damages to the internal dynamics and structural integrity of mobile robots and are allowed to occur. In this case, the forces from the collisions may accelerate the mobile robot in its motion direction, and thus the mobile robot achieves its target task in a shorter time. The comparisons in \cite{collision-allowed} also show that the method based on allowable collisions has better performances than the methods based on unallowable collisions.

Motivated by the above discussion, we address the motion control problem of two mobile robots under allowable collisions in this letter. To handle this problem, the effects of the allowable collisions are analyzed to derive a hybrid model. In particular, we first present the conditions for the collision occurrence, and then analyze how the collisions affect the motion of the mobile robots in terms of momentum and kinetic energy. Finally, based on the hybrid system theory, the robot dynamics and the effects of the collisions are combined together to formulate a collision model into a hybrid system for each mobile robot. We would like to emphasize that the collision model is studied in this letter for the first time.

Due to the allowable collisions and their effects on mobile robots, it is necessary to discuss the task accomplishment of mobile robots. We first verify the necessity of control redesign, and then apply impulsive control techniques to propose a control redesign strategy for mobile robots. Specifically, if the motion direction of the mobile robot is changed after the collision, then the control redesign is necessary. The main idea of the control redesign strategy is to impose an impulse input to the mobile robot after the collision such that the mobile robot moves away from the collision position and the re-collision can be avoided. How to impose such an impulse input is presented in detail, and we show the task accomplishment under the redesigned control strategy. 

The main contribution lies in the following three aspects: (1) for the first time, a hybrid modelling of the occurrence and effects of collisions; (2) a novel control redesign strategy, which is based on the impulsive control techniques and ensures the task accomplishment; (3) a comparative example to show the advantages of the proposed control strategy.

This paper is organized as follows. Section \ref{sec-preliminaries} states the preliminaries and Section \ref{sec-problemformulation} formulates the control problem. The collisions are modeled in Section \ref{sec-collisionmodel}. The control strategy is proposed in Section \ref{sec-controlredesign}. Numerical results are given in Section \ref{sec-simulation}, followed by the conclusion in Section \ref{sec-conclusion}.

%%%%%%%%%%%%%%%%%%%%%%%%%%%%%%%%%%%%%%%%%%%%%%%%%%%%%%%%%%%%%%%%%%%%%%%%%%%%%%%%%%%%%%%%%%%%%%%%%%%%%%%
\section{Preliminaries}
\label{sec-preliminaries}
%%%%%%%%%%%%%%%%%%%%%%%%%%%%%%%%%%%%%%%%%%%%%%%%%%%%%%%%%%%%%%%%%%%%%%%%%%%%%%%%%%%%%%%%%%%%%%%%%%%%%%%

In this section, we present the dynamical systems to be considered and recall some preliminaries on control Lyapunov and barrier functions. We start with the notations used throughout this work.
Let $\mathbb{R}_{+}:=[0,+\infty), \mathbb{R}:=(-\infty,+\infty)$ and $\mathbb{R}^{n}$ denote the $n$-dimensional Euclidean space. $|\cdot|$ denotes the Euclidean norm in $\mathbb{R}^{n}$. Given $x\in\mathbb{R}^{n}$ and $y\in\mathbb{R}^{m}$, $(x,y):=[x^{\top}~y^{\top}]^{\top}$. 
The function $\sgn:\mathbb{R}\rightarrow\{\pm1,0\}$ is defined as follows: for any $x\in\mathbb{R}$, $\sgn(x)=1$, if $x>0$; $\sgn(x)=0$, if $x=0$; $\sgn(x)=-1$, if $x<0$. A continuous function $\alpha:\mathbb{R}_{+}\rightarrow\mathbb{R}_{+}$ is of class $\mathcal{K}$, if it is strictly increasing and $\alpha(0)=0$. A continuous function $\beta:\mathbb{R}\rightarrow\mathbb{R}$ is of class $\overline{\mathcal{K}}$, if it is strictly increasing and $\beta(0)=0$. A function $f:\mathbb{R}^{n}\rightarrow\mathbb{R}^{m}$ is locally Lipschitz, if for any $\varepsilon>0$, there exists $L\in\mathbb{R}$ such that $|f(x)-f(y)|\le L|x-y|$, where $x,y\in\{z\in\mathbb{R}^{n}:|z|\le\varepsilon\}$. Given $A,B\subset\mathbb{R}^{n}$, $B\setminus A:=\{x:x\in B,x\notin A\}$.
	
Consider the following nonlinear control system
\begin{align}
    \label{eqn-1}
    \dot{\xi} = f(\xi)+g(\xi)u,
\end{align}
where $\xi\in\mathbb{R}^{n}$ is the state and $u\in\mathbb{R}^{m}$ is the control input. The functions $f:\mathbb{R}^{n} \rightarrow \mathbb{R}^{n}$ and $g:\mathbb{R}^{n} \rightarrow \mathbb{R}^{n\times m}$ are locally Lipschitz, which implies that given any initial state $\xi_{0}\in\mathbb{R}^{n}$, there exists a unique solution $\xi(t,\xi_{0})$ to the system \eqref{eqn-1}; see \cite[Theorem 3.1]{2001Nonlinear}. Given a control input, the system \eqref{eqn-1} is \emph{globally asymptotically stable} (GAS), if for any $\xi_{0}\in\mathbb{R}^{n}$, $|\xi(t,\xi_{0})|\rightarrow0$ as $t\rightarrow\infty$. The stabilization control is to design a feedback controller such that the system \eqref{eqn-1} is GAS.
	
\begin{definition}[\cite{CBFforU}]
    \label{def-1}
    For the system \eqref{eqn-1}, a continuously differentiable function $V:\mathbb{R}^{n}\rightarrow\mathbb{R}_{+}$ is a \emph{control Lyapunov function} (CLF), if there exists $\alpha\in\mathcal{K}$ such that for all $\xi\in\mathbb{R}^{n}$,
    \begin{align}
	\label{eqn-2}
	\inf\nolimits_{u\in\mathbb{R}^{m}}~
	[L_{f}V(\xi)+L_{g}V(\xi)u]\le -\alpha(V(\xi)),
    \end{align}
    where $L_{f}V(\xi):=\frac{\partial V(\xi)}{\partial \xi}f(\xi)$ and $L_{g}V(\xi):=\frac{\partial V(\xi)}{\partial \xi}g(\xi)$.
\end{definition}
	
A set $\mathcal{S} \subset \mathbb{R}^{n}$ is \emph{forward invariant} for the system \eqref{eqn-1}, if for any $\xi_{0}\in\mathcal{S}$, $\xi(t,\xi_{0}) \in \mathcal{S}$ for all $t\ge0$. If the set $\mathcal{S}$ is forward invariant, then the system \eqref{eqn-1} is \emph{safe} with respect to the set $\mathcal{S}$, which is called a \emph{safe set}. To verify the safety of the system \eqref{eqn-1}, the set $\mathcal{S}$ is associated with a continuously differentiable function $h:\mathbb{R}^{n}\rightarrow\mathbb{R}$, and defined as $\mathcal{S}:=\{\xi\in\mathbb{R}^{n}:h(\xi)\ge0\}$. The safety control is to design a controller such that the system \eqref{eqn-1} is safe. 
	
\begin{definition}[\cite{CBFforU}]
    \label{def-2}
    Given a safe set $\mathcal{S} \subset \mathcal{Q} \subset \mathbb{R}^{n}$ for the system \eqref{eqn-1}, a continuously differentiable function $h:\mathcal{Q}\rightarrow\mathbb{R}$ is a \emph{control barrier function} (CBF), if there exists $\beta \in \overline{\mathcal{K}}$ such that for all $\xi\in\mathcal{Q}$,
    \begin{align} 
	\label{eqn-3}
	\sup\nolimits_{u\in\mathbb{R}^{m}}~ 
	[L_{f}h(\xi)+L_{g}h(\xi)u]\ge-\beta(h(\xi)).
    \end{align}
\end{definition}
	
For the system \eqref{eqn-1}, the stabilization and safety control problems are usually combined together and transformed into the following quadratic programming problem; see also \cite{RobustCBF}.
\begin{align}
    \label{eqn-4}
    \begin{aligned}
        \min\nolimits_{u,\eta}~ & 0.5(u^{\top}u+\rho\eta^{\top}\eta) \\ 
        \text{s.t.}\qquad & \gamma(L_{f}V(\xi)+\alpha(V(\xi))) +L_{g}V(\xi)(u+\eta)\le0,\\
	& L_{f}h(\xi)+\beta(h(\xi)) + L_{g}h(\xi)u \ge 0,
    \end{aligned}
\end{align}	
where $\eta\in\mathbb{R}^{m}$ is a relaxation variable, $\rho>0$ is a design parameter, and the function $\gamma:\mathbb{R}\rightarrow\mathbb{R}$ is defined as follows: for any $x\in\mathbb{R}$,  $\gamma(x)=\sigma_{1}x$ with $\sigma_{1}\ge1$, if $x\ge0$; $\gamma(x)=x$, if $x<0$. Based on Karush–Kuhn–Tucker conditions to solve the optimization problem \eqref{eqn-4}, the controller can be derived to guarantee the stabilization and safety simultaneously. The interested readers can find more details in \cite{RobustCBF}.

%%%%%%%%%%%%%%%%%%%%%%%%%%%%%%%%%%%%%%%%%%%%%%%%%%%%%%%%%%%%%%%%%%%%%%%%%%%%%%%%%%%%%%%%%%%%%%%%%%%%%%%%
\section{Problem Formulation}
\label{sec-problemformulation}
%%%%%%%%%%%%%%%%%%%%%%%%%%%%%%%%%%%%%%%%%%%%%%%%%%%%%%%%%%%%%%%%%%%%%%%%%%%%%%%%%%%%%%%%%%%%%%%%%%%%%%%%
	
In this section, we first describe the workspace of mobile robots, then derive predefined controllers for mobile robots, and finally state the problem to be considered.
	
%-------------------------------------------------------------------------------------------------------
\subsection{Workspace}
\label{subsec-workspace}
%-------------------------------------------------------------------------------------------------------
	
Consider a workspace $\mathbb{S}\subset\mathbb{R}^{2}$ with $n$ cylindrical uniform rigid bodies, where $n\ge3$ is an integer. To facilitate the notation afterwards, let $\mathcal{R}_{1}$ and $\mathcal{R}_{2}$ denote two mobile robots, whereas $\mathcal{R}_{3}, \cdots, \mathcal{R}_{n}$ denote static obstacles. From the properties of uniform rigid bodies \cite[Chapter 8]{CollegePhysics}, the motion of $\mathcal{R}_{\iota}$ can be described by the motion of its centroids, where $\iota\in\{1,\cdots,n\}$. That is, the position of $\mathcal{R}_{\iota}$ is represented by its centroid $p_{\iota}:=(x_{\iota},y_{\iota}) \in\mathbb{S}$. Without loss of generality, the heights of rigid bodies are assumed to be within a reasonable range, and have no effects on the motion. Let $r_{\iota},m_{\iota}>0$ denote the radius and mass of $\mathcal{R}_{\iota}$, respectively. The following assumption is made for all rigid bodies.
	
\begin{assumption}
    \label{ass-1}
    The mass of each obstacle is sufficiently larger than that of any mobile robot.
\end{assumption}
	
Under Assumption \ref{ass-1}, the motion of the obstacles can be ignored, which will be explained in detail in Section \ref{subsec-modellingofRBC}.
	
%-------------------------------------------------------------------------------------------------------
\subsection{Robot Controller}
\label{subsec-robotcontroller}
%-------------------------------------------------------------------------------------------------------
	
Consider two mobile robots with the following dynamics
\begin{align}
    \label{eqn-5}
    \begin{bmatrix}
	\dot{x}_{i} \\ \dot{y}_{i} \\ \dot{\theta}_{i}
    \end{bmatrix}=
    \begin{bmatrix}
        \cos{\theta}_{i} & 0 \\
	\sin{\theta}_{i} & 0 \\ 
	0 & 1
    \end{bmatrix}
    \begin{bmatrix}
	v_{i} \\ w_{i}
    \end{bmatrix} \quad\Rightarrow\quad \dot{\xi}_{i}=g_{i}(\xi_{i})u_{i},
\end{align}
where $i\in\{1,2\},\xi_{i}:=(x_{i},y_{i},\theta_{i}) \in \mathbb{X}:= \mathbb{S}\times\mathbb{R}$ is the state and $u_{i}:=(v_{i},w_{i}) \in \mathbb{R}^{2}$ is the control input. Specifically, $\theta_{i}\in\mathbb{R}$ is the motion direction, $v_{i}\in\mathbb{R}$ is the linear velocity and $w_{i}\in\mathbb{R}$ is the angular velocity. Due to the physical limits, $v_{i}$ and $w_{i}$ are bounded. Therefore, the control input constraint is $u_{i}\in\mathbb{U}:= \mathbb{V}\times\mathbb{W} =\{v_{i}\in\mathbb{R}:|v_{i}|\le M_{v}\}\times\{w_{i}\in\mathbb{R}:|w_{i}|\le M_{w}\}$, where $M_{v},M_{w}>0$ are respectively the maximums of linear and angular velocities.

The tasks of the two mobile robots are to converge to the target states while avoiding collisions. Let $\xi_{id}\in\mathbb{X}$ be the target state of robot $i$. The task of robot $i$ can be divided into two parts. The first part is to converge to the target state $\xi_{id}$, which can be formulated into a stabilization problem. We define the CLF $V_{i}(\xi_{i}) = 0.5|\xi_{i}-\xi_{id}|^{2}$. The second part is the collision avoidance. To this end, we define the CBF $h_{i}(\xi_{i}) = |p_{1}-p_{2}|^{2}-(r_{1}+r_{2})^{2} +\sum_{k=3}^{n}(|p_{i}-p_{k}|^{2}-(r_{i}+r_{k})^{2})$, where $p_{1},p_{2}\in\mathbb{R}^{2}$  are the current positions of the two mobile robots and $r_{1},r_{2}\in\mathbb{R}$ are the radii of the two mobile robots. To achieve these two parts simultaneously and guarantee the satisfaction of the control input constraint, we solve the optimization problem \eqref{eqn-4} with the constraint $u_{i}\in\mathbb{U}$ to obtain the following predefined controller.
\begin{align}
    \label{eqn-6}
    u_{i}^{\textrm{pre}}&=\left\{\begin{aligned}
        &u_{i}^{\textrm{nom}}, &&\text{ if } 
        |v_{i}|\le M_{v},|w_{i}|\le M_{w},\\
        &\left(v_{i}^{\textrm{\textrm{sat}}}, w_{i}^{\textrm{nom}}\right), &&\text{ if } |v_{i}|>M_{v},|w_{i}|\le M_{w},\\
        &\left(v_{i}^{\textrm{nom}}, w_{i}^{\textrm{sat}}\right), &&\text{ if } |v_{i}|\le M_{v},|w_{i}|>M_{w},\\
        &\left(v_{i}^{\textrm{sat}}, w_{i}^{\textrm{sat}}\right), &&\text{ if }|v_{i}|>M_{v},|w_{i}|>M_{w},
    \end{aligned}\right.
\end{align}
where $v_{i}^{\textrm{sat}}=M_{v}\sgn(v_{i}^{\textrm{nom}}), w_{i}^{\textrm{sat}}=M_{w}\sgn(w_{i}^{\textrm{nom}})$ and $u_{i}^{\textrm{nom}}=(v_{i}^{\textrm{nom}},w_{i}^{\textrm{nom}})$ are as follows
\begin{align*}
    u_{i}^{\textrm{nom}}&=\left\{\begin{aligned}
	&(0,0), &&\xi_{i}\in\Omega_{1},\\
        &\tfrac{-\rho}{\rho+1}\left(\tfrac{\mathfrak{a}_{i}(\xi_{i})\mathfrak{c}_{i}(\xi_{i})}{\mathfrak{c}_{i}^{2}(\xi_{i})+\mathfrak{s}_{i}^{2}(\xi_{i})}, \tfrac{\mathfrak{a}_{i}(\xi_{i})\mathfrak{s}_{i}(\xi_{i})}{\mathfrak{c}_{i}^{2}(\xi_{i})+\mathfrak{s}_{i}^{2}(\xi_{i})}\right), &&\xi_{i}\in\Omega_{2},\\
        &\left(\tfrac{-\mathfrak{b}_{i}(\xi_{i})}{\mathfrak{e}_{i}(\xi_{i})},0\right), &&\xi_{i}\in\Omega_{3},\\
        &\left(\tfrac{-\mathfrak{b}_{i}(\xi_{i})}{\mathfrak{e}_{i}(\xi_{i})},\tfrac{\mathfrak{b}_{i}(\xi_{i})\mathfrak{c}_{i}(\xi_{i})-\mathfrak{a}_{i}(\xi_{i})\mathfrak{e}_{i}(\xi_{i})}{\frac{1}{\rho}\mathfrak{c}_{i}^{2}(\xi_{i})+\frac{\rho+1}{\rho}\mathfrak{s}_{i}^{2}(\xi_{i})}\tfrac{\mathfrak{s}_{i}(\xi_{i})}{\mathfrak{e}_{i}(\xi_{i})}\right), &&\xi_{i}\in\Omega_{4},
    \end{aligned}\right.
\end{align*}
where $\rho>0, \mathfrak{a}_{i}(\xi_{i})=\gamma(\sigma_{2}V_{i}(\xi_{i})), \mathfrak{b}_{i}(\xi_{i})=\sigma_{3}h_{i}(\xi_{i})$ with $\sigma_{2},\sigma_{3}>0$, $(\mathfrak{c}_{i}(\xi_{i}), \mathfrak{s}_{i}(\xi_{i}))=(L_{g_{i}}V_{i}(\xi_{i}))^{\top}, (\mathfrak{e}_{i}(\xi_{i}),0)=(L_{g_{i}}h_{i}(\xi_{i}))^{\top}$, $\Omega_{1}=\{\xi_{i}\in\mathbb{X}: \mathfrak{a}_{i}(\xi_{i})<0, \mathfrak{b}_{i}(\xi_{i})>0\}, 
\Omega_{2}=\{\xi_{i}\in\mathbb{X}: \mathfrak{a}_{i}(\xi_{i})\ge0, \mathfrak{b}_{i}(\xi_{i})>\frac{\rho\mathfrak{e}_{i}(\xi_{i})\mathfrak{c}_{i}(\xi_{i})\mathfrak{a}_{i}(\xi_{i})}{(\rho+1)(\mathfrak{c}_{i}^{2}(\xi_{i})+\mathfrak{s}_{i}^{2}(\xi_{i}))}\}, \Omega_{3}=\{\xi_{i}\in\mathbb{X}: \mathfrak{a}_{i}(\xi_{i})<\frac{\mathfrak{c}_{i}(\xi_{i})\mathfrak{b}_{i}(\xi_{i})}{\mathfrak{e}_{i}(\xi_{i})}, \mathfrak{b}_{i}(\xi_{i})\le0\}$ and $\Omega_{4}=\{\xi_{i}\in\mathbb{X}\backslash\Omega_{1}: \mathfrak{a}_{i}(\xi_{i})\ge\frac{\mathfrak{c}_{i}(\xi_{i})\mathfrak{b}_{i}(\xi_{i})}{\mathfrak{e}_{i}(\xi_{i})}, \mathfrak{b}_{i}(\xi_{i})\le\frac{\rho \mathfrak{e}_{i}(\xi_{i})\mathfrak{c}_{i}(\xi_{i}) \mathfrak{a}_{i}(\xi_{i})}{(\rho+1)(\mathfrak{c}_{i}^{2}(\xi_{i})+\mathfrak{s}_{i}^{2}(\xi_{i}))}\}$.
	
%-------------------------------------------------------------------------------------------------------
\subsection{Problem Statement}
\label{subsec-problemformulation}
%-------------------------------------------------------------------------------------------------------
	
For the mobile robot with the predefined controller \eqref{eqn-6}, the task accomplishment may not be ensured. To show this, the following example is presented.
	
\begin{figure}[!t]
    \begin{center}
	\begin{picture}(45, 110)
            \put(-65, -25){\resizebox{63mm}{46mm}{\includegraphics[width=2.5in]{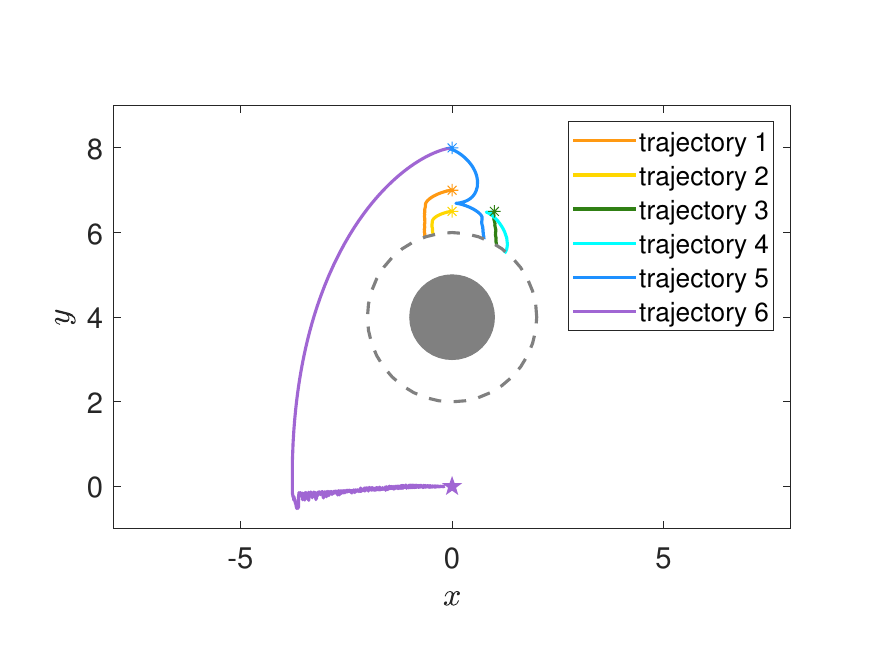}}}
	\end{picture}
    \end{center}
    \caption{Illustration of the trajectories starting from different initial states.}
    \label{fig-1}
\end{figure}
	
\begin{example}
    \label{example-1}
    Consider the workspace $\mathbb{S}=[-8,8] \times [-1,9]$, where a static obstacle is centered at $p_{k}=(0,4)$ with the radius $r_{k}=1$. A mobile robot with the radius $r=1$ is of the dynamics \eqref{eqn-5}, and its target state is $\xi_{d}=(0, 0, 0.5\pi)$. To reach the target state $\xi_{d}$, the CLF and the CBF are defined as $V(\xi)=0.5|\xi-\xi_{d}|^{2}$ and $h(\xi)=|p-p_{k}|^{2}-4$, respectively. Let $\rho=9, \sigma_{1}=1.25, \sigma_{2}=0.6, \sigma_{3}=1.2$ and $M_{v}=M_{w}=5$. With the predefined controller \eqref{eqn-6}, the trajectory from any initial state can be derived. Given the initial states $\xi^{1}_{0}=(0,7,0.01\pi), \xi^{2}_{0}=(0,6.5,0.01\pi), \xi^{3}_{0}=(1,6.5,0.3\pi), \xi^{4}_{0}=(1,6.5,2.1\pi), \xi^{5}_{0}=(0,8,2.1\pi)$ and $\xi^{6}_{0}=(0,8,0.01\pi)$, the corresponding trajectories are shown in Fig. \ref{fig-1}, in which only the trajectory from $\xi^{6}_{0}$ reaches the target state. From $\xi^{5}_{0}$ and $\xi^{6}_{0}$ ($\xi^{3}_{0}$ and $\xi^{4}_{0}$), we see the effects of the initial motion directions on the robot motion and further on the task accomplishment. Similarly, from $\xi^{1}_{0}, \xi^{2}_{0}$ and $\xi^{6}_{0}$, we see the effects of the initial positions on the task accomplishment. $\hfill\diamondsuit$
\end{example}
	
In Example \ref{example-1}, the CBF has been used to derive the predefined controller, while the collisions do still exist in Fig. \ref{fig-1}. We conclude that the CBF cannot always ensure the collision avoidance. If multiple robots move simultaneously in a common workspace, then the inter-robot collisions are inevitable. Let mobile robots and obstacles be the rigid bodies. From the properties of rigid bodies \cite[Chapter 7]{CollegePhysics}, each point in a rigid body is in the same position relative to other points, even if the external forces are acted on the rigid body. That is, the collisions do not affect the structure of the rigid body. In these respects, the unavoidable collisions are allowed to occur in practical applications. For these collisions, the following assumption is made.

\begin{figure}
    \begin{center}
	\begin{picture}(40, 70)
            \put(-25, -15){\resizebox{33mm}{21mm}{\includegraphics[width=2.5in]{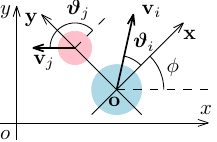}}}
	\end{picture}
    \end{center}
    \caption{Illustration of the construction of the local coordinate frame.}
    \label{fig-2}
\end{figure}

\begin{assumption}
    \label{ass-2}
    For each robot, if the collision occurs, then one and only one collision occurs at each time instant.
\end{assumption}

From Assumption \ref{ass-2}, more than two collisions cannot occur at the same time, and thus both chattering and deadlock phenomena caused by simultaneous multiple collisions are excluded. Under Assumptions \ref{ass-1}-\ref{ass-2}, the problem to be considered is stated below.

\begin{problem}
    \label{problem-1}
    Consider two mobile robots and finite static obstacles in a common workspace. Each robot has its own task and predefined controller \eqref{eqn-6}. 
    First, determine whether the collisions occur. Second, if the collisions occur, then determine the effects of the collisions on the robot motion, and propose a control redesign strategy for each mobile robot to guarantee the accomplishment of the motion control task.
\end{problem}
	
To solve Problem \ref{problem-1}, we first formulate the collision model into a hybrid system (see \cite{HybridDS} for more details) in Section \ref{sec-collisionmodel}, and then redesign the control strategy in Section \ref{sec-controlredesign}.

%%%%%%%%%%%%%%%%%%%%%%%%%%%%%%%%%%%%%%%%%%%%%%%%%%%%%%%%%%%%%%%%%%%%%%%%%%%%%%%%%%%%%%%%%%%%%%%%%%%%%%%
\section{Collision Model}
\label{sec-collisionmodel}
%%%%%%%%%%%%%%%%%%%%%%%%%%%%%%%%%%%%%%%%%%%%%%%%%%%%%%%%%%%%%%%%%%%%%%%%%%%%%%%%%%%%%%%%%%%%%%%%%%%%%%%

In this section, we first derive the condition for the collision occurrence, and then investigate the effects of the collisions on the robot motion to formulate a hybrid model.

%-------------------------------------------------------------------------------------------------------
\subsection{Collision Between Two Rigid Bodies}
\label{subsec-collisionbetweentworigidbodies}
%-------------------------------------------------------------------------------------------------------

The robot model \eqref{eqn-5} is built in the \textit{global coordinate frame} ($gcf$). Note that the collision between $\mathcal{R}_{i}$ and $\mathcal{R}_{j}$ depends on their relative position, where $i\in\{1,2\}$ and $j\in\{1,\cdots,n\}\backslash\{i\}$. We construct a \textit{local coordinate frame} ($lcf$) via the following three steps, as shown in Fig. \ref{fig-2}. 
\begin{enumerate}[a)]
	\item The origin of the $lcf$ is the centroid of $\mathcal{R}_{i}$, and its position in the $gcf$ is denoted as $(x_{i\mathbf{o}}^{j},y_{i\mathbf{o}}^{j})$.
	
	\item The centroids of $\mathcal{R}_{i}$ and $\mathcal{R}_{j}$ are connected as the $\mathbf{y}$-axis of the $lcf$, and the direction from $\mathcal{R}_{i}$ to $\mathcal{R}_{j}$ is the positive direction of the $\mathbf{y}$-axis.
	
	\item The $\mathbf{x}$-axis of the $lcf$ is the vertical of the $\mathbf{y}$-axis. The positive direction of the $\mathbf{x}$-axis is the direction of rotating the positive direction of the $\mathbf{y}$-axis clockwise by $0.5\pi$.
\end{enumerate}
	
Let $\phi_{i}^{j}\in[0,2\pi)$ be the angle between the $x$- and $\mathbf{x}$-axes. Based on the rotation and translation techniques, the $lcf$ can be obtained from the $gcf$, and thus the state in the $gcf$ can be transformed to the state in the $lcf$, which is shown below. 
\begin{align}
    \label{eqn-7}
    \underbrace{\begin{bmatrix}
	\mathbf{x}_{i} \\ \mathbf{y}_{i} \\ 
	\bm{\vartheta}_{i} \end{bmatrix}}_{\bm{\zeta}_{i}} =
    \underbrace{\begin{bmatrix}
	\cos\phi_{i}^{j} & \sin\phi_{i}^{j} & 0 \\
	-\sin\phi_{i}^{j} & \cos\phi_{i}^{j} & 0 \\
	0 & 0 & 1 \end{bmatrix}}_{T_{i}^{j}}\left(\begin{bmatrix}
	x_{i} \\ y_{i} \\ \theta_{i}
    \end{bmatrix}-\begin{bmatrix}
        x_{i\mathbf{o}}^{j} \\ y_{i\mathbf{o}}^{j} \\ \phi_{i}^{j}
    \end{bmatrix}\right),
\end{align}
where $\xi_{i}$ is the global state, $\bm{\zeta}_{i}\in\mathbb{S}^{\prime}\times\mathbb{R}$ is the local state and $\mathbb{S}^{\prime}\subset\mathbb{R}^{2}$ is the workspace in the $lcf$. Since the matrix $T_{i}^{j}$ is invertible, the above transformation is reversible. In addition, the linear and angular velocities are scalars, and thus their values are independent of the reference coordinate frame.
\begin{align}
    \label{eqn-8}
    \mathbf{v}_{i} = v_{i}, \quad 
    \mathbf{w}_{i} = w_{i},
\end{align}
where $\mathbf{v}_{i}\in\mathbb{V}, \mathbf{w}_{i}\in\mathbb{W}$ are respectively the local linear and angular velocities. In the $lcf$, we decompose $\mathbf{v}_{i}$ as follows.
\begin{align}
    \label{eqn-9}
    \mathbf{v}_{i\mathbf{x}} = \mathbf{v}_{i}\cos\bm{\vartheta}_{i}, \quad \mathbf{v}_{i\mathbf{y}} = \mathbf{v}_{i}\sin\bm{\vartheta}_{i},
\end{align}
where $\mathbf{v}_{i\mathbf{x}}, \mathbf{v}_{i\mathbf{y}}\in\mathbb{V}$ are the local linear velocities along the $\mathbf{x}$- and $\mathbf{y}$-axes, respectively. The transformation in \eqref{eqn-7}-\eqref{eqn-9} is available for $\mathcal{R}_{j}$. In particular, if $\mathcal{R}_{j}$ is static, then $\theta_{j}=v_{j}=w_{j}=0$. In the $lcf$, the definition of the collision between two rigid bodies is given as follows.
	
\begin{definition}
    \label{def-3}
    For two rigid bodies, an event between them is called a \emph{rigid body collision} (RBC), if
    \begin{enumerate}[1)]
        \item their distance equals to the sum of their radii;
        \item at least one local linear velocity changes instantaneously.
    \end{enumerate}
\end{definition}

In Definition \ref{def-3}, item 1) is the distance condition for the collision occurrence, which means that two rigid bodies meet each other, and item 2) shows the effect of the collision, which is the abrupt change of the local linear velocity of at least one rigid body. If only item 1) holds, then the two rigid bodies just meet each other and have no mutual effects.

Since $\mathcal{R}_{i}$ is centered at the origin of the $lcf$, the distance between $\mathcal{R}_{i}$ and $\mathcal{R}_{j}$ is $\mathbf{d}_{ij}:=|\mathbf{p}_{j}|$, where $\mathbf{p}_{j}=(\mathbf{x}_{j},\mathbf{y}_{j})\in\mathbb{S}^{\prime}$ is the local position of $\mathcal{R}_{j}$. The following theorem states the condition for the occurrence of the RBC.
	
\begin{theorem}
    \label{thm-1}
    Consider two rigid bodies with their distance $\mathbf{d}_{ij}$ and local linear velocities $\mathbf{v}_{i},\mathbf{v}_{j}$. If $\mathbf{d}_{ij}=r_{i}+r_{j}$ and $\mathbf{v}_{i\mathbf{y}}>\mathbf{v}_{j\mathbf{y}}$, then the RBC occurs.
\end{theorem}

\begin{IEEEproof}
	Since $\mathbf{d}_{ij}=r_{i}+r_{j}$, item 1) in Definition \ref{def-3} holds. From \eqref{eqn-9}, $\mathbf{v}_{i}$ and $\mathbf{v}_{j}$ can be decomposed along the $\mathbf{x}$- and $\mathbf{y}$-axes. Due to the construction of the $lcf$, there is no interactive force between the two rigid bodies in the $\mathbf{x}$-axis, and thus the local linear velocities along the $\mathbf{x}$-axis are not changed. If $\mathbf{v}_{i\mathbf{y}} > \mathbf{v}_{j\mathbf{y}}$, then there exist interactive forces between the two rigid bodies in the $\mathbf{y}$-axis. Hence, the local linear velocities along the $\mathbf{y}$-axis are changed, which further implies the changes of the local linear velocities. That is, item 2) in Definition \ref{def-3} holds. Therefore, we claim that the RBC occurs and the proof is completed.
\end{IEEEproof}

From Theorem \ref{thm-1}, the next corollary is derived directly to show the conditions for the non-occurrence of the RBC. 

\begin{corollary}
    \label{cor-1}
    Consider two rigid bodies with their distance $\mathbf{d}_{ij}$ and local linear velocities $\mathbf{v}_{i},\mathbf{v}_{j}$. If $\mathbf{d}_{ij}=r_{i}+r_{j}$ and $\mathbf{v}_{i\mathbf{y}} \le \mathbf{v}_{j\mathbf{y}}$ (or $\mathbf{d}_{ij} > r_{i}+r_{j}$), then no RBC occurs.
\end{corollary}
	
%-------------------------------------------------------------------------------------------------------
\subsection{Modelling of Rigid Body Collision}
\label{subsec-modellingofRBC}
%-------------------------------------------------------------------------------------------------------
	
After analyzing the occurrence of the RBC, we investigate the effects model of the RBC in this subsection. Since the RBC occurs instantaneously, the positions of two rigid bodies are not changed after the collision. That is,
\begin{align}
    \label{eqn-10}
    \mathbf{x}_{i}^{+} = \mathbf{x}_{i}, \quad \mathbf{y}_{i}^{+} = \mathbf{y}_{i}, \quad \mathbf{x}_{j}^{+} = \mathbf{x}_{j}, \quad \mathbf{y}_{j}^{+} = \mathbf{y}_{j},
\end{align}
where $(\mathbf{x}_{i}^{+},\mathbf{y}_{i}^{+}),(\mathbf{x}_{j}^{+},\mathbf{y}_{j}^{+})\in\mathbb{S}^{\prime}$ are the positions of $\mathcal{R}_{i}$ and $\mathcal{R}_{j}$ after the RBC, $i\in\{1,2\}$ and $j\in\{1,\cdots,n\}\backslash\{i\}$. In terms of energy consumption, the following assumption is made to facilitate the analysis afterwards.
	
\begin{assumption}
    \label{ass-3}
    There is no energy loss in each RBC.
\end{assumption}

If the two rigid bodies involved in an RBC are treated as a system, then Assumption \ref{ass-3} implies the conservation of both momentum and kinetic energy in this system \cite[Chapter 6]{CollegePhysics}. In this way, we next consider the $\mathbf{x}$-axis case and the $\mathbf{y}$-axis case. Since there is no interactive force between $\mathcal{R}_{i}$ and $\mathcal{R}_{j}$ in the $\mathbf{x}$-axis, the local linear velocities along the $\mathbf{x}$-axis after the RBC are not changed. That is, $\mathbf{v}_{i\mathbf{x}}^{+} = \mathbf{v}_{i\mathbf{x}}$ and $\mathbf{v}_{j\mathbf{x}}^{+} = \mathbf{v}_{j\mathbf{x}}$, where $\mathbf{v}_{i\mathbf{x}}^{+},\mathbf{v}_{j\mathbf{x}}^{+}\in\mathbb{V}$ are the local linear velocities along the $\mathbf{x}$-axis after the RBC. Therefore, both momentum and kinetic energy in the $\mathbf{x}$-axis are conserved. Based on Assumption \ref{ass-3}, both momentum and kinetic energy in the $\mathbf{y}$-axis are also conserved. The conversation in the $\mathbf{y}$-axis can be written as
\begin{align}
    \label{eqn-11}
    \begin{aligned}
        m_{i}\mathbf{v}_{i\mathbf{y}} + m_{j}\mathbf{v}_{j\mathbf{y}} & = m_{i}\mathbf{v}_{i\mathbf{y}}^{+} + m_{j}\mathbf{v}_{j\mathbf{y}}^{+},  \\
        \tfrac{1}{2}m_{i}\mathbf{v}_{i\mathbf{y}}^{2} +\tfrac{1}{2}m_{j}\mathbf{v}_{j\mathbf{y}}^{2} & = \tfrac{1}{2}m_{i}(\mathbf{v}_{i\mathbf{y}}^{+})^{2} +\tfrac{1}{2}m_{j}(\mathbf{v}_{j\mathbf{y}}^{+})^{2},
    \end{aligned}
\end{align}
where $\mathbf{v}_{i\mathbf{y}}^{+},\mathbf{v}_{j\mathbf{y}}^{+}\in\mathbb{V}$ are the local linear velocities along the $\mathbf{y}$-axis after the RBC. From \eqref{eqn-11}, we obtain the jump of the local linear velocity
\begin{align}  
    \label{eqn-12}
    \mathbf{v}_{i}^{+} = \left((\mathbf{v}_{i\mathbf{y}}^{+})^{2}+(\mathbf{v}_{i\mathbf{x}}^{+})^{2}\right)^{\frac{1}{2}}\sgn(\mathbf{v}_{i\mathbf{y}}^{+}),
\end{align}
where $\mathbf{v}_{i\mathbf{y}}^{+} = \frac{m_{i}-m_{j}}{m_{i}+m_{j}}\mathbf{v}_{i\mathbf{y}}+\frac{2m_{j}}{m_{i}+m_{j}}\mathbf{v}_{j\mathbf{y}}$ and $\mathbf{v}_{i\mathbf{x}}^{+} = \mathbf{v}_{i\mathbf{x}}$. The local motion direction is updated as 
\begin{align}
    \label{eqn-13}
    \bm{\vartheta}_{i}^{+}=\left\{\begin{aligned}
        &0.5\pi\sgn(\mathbf{v}_{i\mathbf{y}}^{+}), &&\text{if } \mathbf{v}_{i\mathbf{x}}^{+}=0,\mathbf{v}_{i\mathbf{y}}^{+}\in\mathbb{V}, \\
        &0.5\pi\left(1-\sgn(\mathbf{v}_{i\mathbf{x}}^{+})\right), &&\text{if } \mathbf{v}_{i\mathbf{x}}^{+}\ne0, \mathbf{v}_{i\mathbf{y}}^{+}=0, \\
        &\arctan({\mathbf{v}_{i\mathbf{y}}^{+}}/{\mathbf{v}_{i\mathbf{x}}^{+}}), &&\text{otherwise}.
    \end{aligned}\right.
\end{align}
Similarly, we derive $\mathbf{v}_{j}^{+}$ and $\bm{\vartheta}_{j}^{+}$ in the same fashion.
	
\begin{remark}
\label{rmk-1}
Without Assumption \ref{ass-3}, we need to consider the energy loss. In this case, the coefficient $\delta_{i}\in(0,1)$ is introduced such that $\mathbf{v}_{i\mathbf{y}}^{+} = (1-\delta_{i}) (\tfrac{m_{i}-m_{j}}{m_{i}+m_{j}} \mathbf{v}_{i\mathbf{y}} $ $+ \tfrac{2m_{j}}{m_{i}+m_{j}} \mathbf{v}_{j\mathbf{y}})$, where $1-\delta_{i}$ indicates the remaining velocity after the RBC. The following analysis is still valid and is omitted here. $\hfill\square$
\end{remark}

In the $lcf$, there exist interactive forces in the $\mathbf{y}$-axis only and the centroids of $\mathcal{R}_{i}$ and $\mathcal{R}_{j}$ are on the $\mathbf{y}$-axis. Since $\mathcal{R}_{i}$ and $\mathcal{R}_{j}$ are assumed to be the cylindrical uniform rigid bodies, their centroids are on their rotation axes \cite[Chapter 8]{CollegePhysics}. Hence, the forces from the RBC intersect with the rotation axes of two rigid bodies at their centroids. That is, the forces from the RBC and their arms are in the same line, which means the zero torque caused by the RBC \cite[Chapter 8]{CollegePhysics}. Therefore, the local angular velocities of $\mathcal{R}_{i}$ and $\mathcal{R}_{j}$ are not changed after the RBC. That is, $\mathbf{w}_{i}^{+}=\mathbf{w}_{i}$ and $\mathbf{w}_{j}^{+}=\mathbf{w}_{j}$.

From the above analysis, we obtain
    \begin{align}
	\label{eqn-14}
        (\mathbf{x}_{i}^{+}, \mathbf{y}_{i}^{+}, \bm{\vartheta}_{i}^{+}, \mathbf{v}_{i}^{+}, \mathbf{w}_{i}^{+}) &=(\mathbf{x}_{i}, \mathbf{y}_{i}, \bm{\vartheta}_{i}^{+}, \mathbf{v}_{i}^{+}, \mathbf{w}_{i}).
    \end{align}
Due to \eqref{eqn-14} and the construction of the $lcf$, the $lcf$ is invariant after the RBC. That is, the angle $\phi_{i}^{j}$ is not changed. Combining \eqref{eqn-7}-\eqref{eqn-8} and \eqref{eqn-14} obtains the jumps of the state, linear and angular velocities in the $gcf$ as follows.
\begin{subequations}
    \label{eqn-15}
    \begin{align}
        (x_{i}^{+}, y_{i}^{+}, \theta_{i}^{+}) &= (T_{i}^{j})^{-1}(\mathbf{x}_{i}^{+},\mathbf{y}_{i}^{+},\bm{\vartheta}_{i}^{+})+(x_{i\mathbf{o}}^{j},y_{i\mathbf{o}}^{j},\phi_{i}^{j}) \nonumber \\
	\label{eqn-15a} 
        &=(x_{i},y_{i},\theta_{i}^{+}),\\
	\label{eqn-15b}
        (v_{i}^{+},w_{i}^{+}) &=(\mathbf{v}_{i}^{+},\mathbf{w}_{i}^{+})=(\mathbf{v}_{i}^{+},\mathbf{w}_{i}) =(v_{i}^{+},w_{i}).
	\end{align}
\end{subequations}
The variables $\theta_{i}^{+}$ and $v_{i}^{+}$ in \eqref{eqn-15} are derived below. Based on \eqref{eqn-7}-\eqref{eqn-9}, we define $\bar{\theta}_{i}=\theta_{i}-\phi_{i}^{j}$ and obtain
\begin{align}
    \label{eqn-16}
    \mathbf{v}_{i\mathbf{x}} = v_{i}\cos\bar{\theta}_{i},\quad \mathbf{v}_{i\mathbf{y}} = v_{i}\sin\bar{\theta}_{i}.
\end{align}
Substituting \eqref{eqn-16} into \eqref{eqn-12}-\eqref{eqn-13} yields the jumps of the global motion direction and the global linear velocity below.
\begin{align}
    \label{eqn-17}
    \theta_{i}^{+}&=\left\{\begin{aligned}
        &0.5\pi\sgn\left(\lambda_{i}\right),\quad &&\text{if } \mu_{i}=0,\lambda_{i}\in\mathbb{V}, \\
        &0.5\pi\left(1-\sgn(\mu_{i})\right), \quad &&\text{if } \mu_{i}\ne0,\lambda_{i}=0, \\
        &\arctan(\lambda_{i}/\mu_{i}) +\phi,\quad &&\text{otherwise},
    \end{aligned}\right. \\    
    \label{eqn-18}
    v_{i}^{+}&=\left(\lambda_{i}^{2}+\mu_{i}^{2}\right)^{\frac{1}{2}}\sgn(\lambda_{i}),
\end{align}	
where $\lambda_{i}=\frac{m_{i}-m_{j}}{m_{i}+m_{j}}v_{i}\sin\bar{\theta}_{i}+\frac{2m_{j}}{m_{i}+m_{j}}v_{j}\sin\bar{\theta}_{j}$ and $\mu_{i}=v_{i}\cos\bar{\theta}_{i}$. Following the similar fashion, we derive the updates for $\mathcal{R}_{j}$ after the RBC.
	
From \eqref{eqn-15}, the RBC has no effects on the positions and angular velocities of the two rigid bodies. Hence, next we only study the effects of the RBC on the motion directions and linear velocities. To show this, two special cases of the RBC between $\mathcal{R}_{i}$ and $\mathcal{R}_{j}$ are discussed below.

\subsubsection*{\textbf{Case 1}} 
\emph{$\mathcal{R}_{i}$ and $\mathcal{R}_{j}$ move in the same or opposite direction.} In this case, $v_{i}\cos\bar{\theta}_{i}=v_{j}\cos\bar{\theta}_{j}=0$. Thus, $\lambda_{i}=\frac{m_{i}-m_{j}}{m_{i}+m_{j}}v_{i}\sin\bar{\theta}_{i}+\frac{2m_{j}}{m_{i}+m_{j}}v_{j}\sin\bar{\theta}_{j},\lambda_{j}=\frac{m_{j}-m_{i}}{m_{i}+m_{j}}v_{j}\sin\bar{\theta}_{j}+\frac{2m_{i}}{m_{i}+m_{j}}v_{i}\sin\bar{\theta}_{i}$ and $\mu_{i}=\mu_{j}=0$. From \eqref{eqn-17}, if $\sgn(\lambda_{i})=\sgn(v_{i}\sin\bar{\theta}_{i})$, then $\theta_{i}^{+}=\theta_{i}$; otherwise, $\theta_{i}^{+}\ne\theta_{i}$. Similar results are derived for $\mathcal{R}_{j}$. From \eqref{eqn-18}, the linear velocities of $\mathcal{R}_{i}$ and $\mathcal{R}_{j}$ are changed. Hence, $\mathcal{R}_{i}$ and $\mathcal{R}_{j}$ may move in the same direction at different speeds after the RBC.

\subsubsection*{\textbf{Case 2}}
\emph{$\mathcal{R}_{j}$ is static and has a sufficiently large mass.}
In this case, $v_{j}=0, v_{i}\ne0, \frac{m_{i}-m_{j}}{m_{i}+m_{j}}\approx-1$ and $\frac{2m_{i}}{m_{i}+m_{j}}\approx0$. That is, $\lambda_{i}=-v_{i}\sin\bar{\theta}_{i},\mu_{i}=v_{i}\cos\bar{\theta}_{i}$ and $\lambda_{j}=\mu_{j}=0$.
From \eqref{eqn-17}-\eqref{eqn-18}, $\theta_{i}^{+}\ne\theta_{i}, \theta_{j}^{+}=\theta_{j}, v_{i}^{+}=v_{i}$ and $v_{j}^{+}=0$. Hence, $\mathcal{R}_{i}$ moves in a different direction at the same speed after the RBC, whereas $\mathcal{R}_{j}$ is still static.
	
From Assumption \ref{ass-1} and Case 2, all static obstacles in the workspace are still static after the RBCs. That is, the RBCs have no effects on the obstacles and the motion of the obstacles can be ignored. Hence, we only address the effects of the RBCs on the mobile robots. Based on Theorem \ref{thm-1}, Corollary \ref{cor-1} and \eqref{eqn-16}, we define the following two sets: $\mathcal{C}_{i}=\{\xi_{i}\in\mathbb{X},u_{i}\in\mathbb{U}:d_{ij}>r_{i}+r_{j}\}\cup\{\xi_{i}\in\mathbb{X},u_{i}\in\mathbb{U}:d_{ij}=r_{i}+r_{j},v_{i}\sin\bar{\theta}_{i} \le v_{j}\sin\bar{\theta}_{j}\}$ and $\mathcal{D}_{i}=\{\xi_{i}\in\mathbb{X},u_{i}\in\mathbb{U}:d_{ij}=r_{i}+r_{j},v_{i}\sin\bar{\theta}_{i}>v_{j}\sin\bar{\theta}_{j}\}$, where $d_{ij}=|p_{i}-p_{j}|$. We can check that the sets $\mathcal{C}_{i}$ and $\mathcal{D}_{i}$ are closed. If $(\xi_{i},u_{i})\in \mathcal{C}_{i}$, then no RBC occurs. If $(\xi_{i},u_{i})\in\mathcal{D}_{i}$, then robot $i$ collides with $\mathcal{R}_{j}$ and $\xi_{i}$ will jump as in \eqref{eqn-15a}. For robot $i$, combining the dynamics \eqref{eqn-5} and the jump \eqref{eqn-15a} formulates the collision model into the hybrid system as follows.
\begin{subequations}
    \label{eqn-19}
    \begin{align}
	    \label{eqn-19a}
        \dot{\xi}_{i} &= \mathfrak{f}_{i}(\xi_{i},u_{i}) :=g_{i}(\xi_{i})u_{i}, && \quad (\xi_{i},u_{i}) \in \mathcal{C}_{i}, \\
	    \label{eqn-19b}
        \xi_{i}^{+} &= \mathfrak{j}_{i} (\xi_{i},u_{i}):=(x_{i},y_{i},\theta_{i}^{+}), && \quad (\xi_{i},u_{i}) \in \mathcal{D}_{i},
    \end{align}
\end{subequations}
where $\mathcal{C}_{i}$ is the flow set, $\mathfrak{f}_{i}:\mathcal{C}_{i}\rightarrow\mathbb{R}^{3}$ is the flow map, $\mathcal{D}_{i}$ is the jump set and $\mathfrak{j}_{i}:\mathcal{D}_{i}\rightarrow\mathbb{X}$ is the jump map. The flow map $\mathfrak{f}_{i}$ and the jump map $\mathfrak{j}_{i}$ are continuous.

%%%%%%%%%%%%%%%%%%%%%%%%%%%%%%%%%%%%%%%%%%%%%%%%%%%%%%%%%%%%%%%%%%%%%%%%%%%%%%%%%%%%%%%%%%%%%%%%%%%%%%%
\section{Control Redesign}
\label{sec-controlredesign}
%%%%%%%%%%%%%%%%%%%%%%%%%%%%%%%%%%%%%%%%%%%%%%%%%%%%%%%%%%%%%%%%%%%%%%%%%%%%%%%%%%%%%%%%%%%%%%%%%%%%%%%

In this section, we start with the necessity of the motion control redesign, then redesign the motion control strategy via impulsive control techniques, and finally show the task accomplishment under the redesigned control strategy.

%-------------------------------------------------------------------------------------------------------
\subsection{Necessity and Strategy of Control Redesign}
\label{subsec-necessityandstrategyofcontrolredesign}
%-------------------------------------------------------------------------------------------------------

From \eqref{eqn-19}, we observe that the RBC affects the motion direction of mobile robots. Hence, we check whether the motion direction is changed after the RBC, and further determine if the predefined controller \eqref{eqn-6} is still valid and the control redesign is needed. If $\theta_{i}^{+}=\theta_{i}$, corresponding to Case 1 in Section \ref{subsec-modellingofRBC}, then the state of robot $i$ is not changed after the RBC. In this case, the predefined controller $u_{i}^{\textrm{pre}}$ is not impacted and thus the control redesign is not needed for robot $i$. If $\theta_{i}^{+}\ne\theta_{i}$, then the state of robot $i$ is changed after the RBC, which implies that the predefined controller $u_{i}^{\textrm{pre}}$ is impacted. In this case, the task accomplishment may be affected, and thus it is necessary to redesign the controller for robot $i$. From the above analysis, we conclude that, for robot $i$ after the RBC, if $\theta_{i}^{+}\ne\theta_{i}$, then its controller needs to be redesigned; otherwise, its controller does not need to be redesigned.

Next, we focus on the case where the control redesign is needed and show how to redesign the controller via impulsive control techniques. Without loss of generality, the RBC between robot $i$ and $\mathcal{R}_{j}$ is considered, where $i\in\{1,2\}$ and $j\in\{1,\cdots,n\}\backslash\{i\}$. Let $p_{ic}^{j}:=(x_{ic}^{j},y_{ic}^{j}) \in\mathbb{S}$ be the position of the collision occurrence, where $|p_{ic}^{j}-p_{j}|=r_{i}+r_{j}$ and $r_{i},r_{j}>0$ are respectively the radii of robot $i$ and $\mathcal{R}_{j}$. For the controller redesign, we aim to avoid the re-occurrence of the collision with $\mathcal{R}_{j}$ and to ensure the task accomplishment. To this end, the following control redesign strategy is proposed.
\begin{enumerate}[a)]
    \item After the RBC, change the motion direction of robot $i$ immediately by imposing an impulse $\xi_{i}^{j}\in\mathbb{R}^{3}$.
	
    \item A local controller $u_{i}^{j}\in\mathbb{U}$ is activated to drive robot $i$ away from the collision position $p_{ic}^{j}$.
	
    \item The predefined controller $u_{i}^{\textrm{pre}}$ in \eqref{eqn-6} is re-activated once the position of robot $i$ satisfies the following conditions:
    \begin{subequations}
	\label{eqn-20}
	\begin{align}
		\label{eqn-20a}
        |p_{i}-p_{\mathrm{k}}|&> r_{i}+r_{\mathrm{k}},\quad\mathrm{k}\in\{1,\cdots,n\}\setminus\{i\}, \\
		\label{eqn-20b}
        |p_{i}-p_{ic}^{j}| &= \left\{\begin{aligned}
            &0.5r_{j}, &&\text{if }j\in\{1,2\}\text{ and }j\ne i, \\
			&r_{j}, &&\text{if }j\in\{3,\cdots,n\}.
		\end{aligned}\right.
	\end{align}
    \end{subequations}
\end{enumerate}

In the proposed three-step strategy, the first two steps are to drive the mobile robot away from the collision position to avoid re-collision, and the last step is to adopt the predefined controller to ensure the task accomplishment. Hence, the first two steps are the key of the control redesign strategy. How to impose the impulse $\xi_{i}^{j}$ and how to design the local controller $u_{i}^{j}$ will be presented in detail in the next subsection.
	
\begin{remark}
    \label{rek-2}
    In the last step of the control redesign strategy, \eqref{eqn-20} shows where to switch the local controller to the predefined controller.  In particular, \eqref{eqn-20a} is the condition for the distance between robot $i$ and any rigid body. \eqref{eqn-20b} is the condition for the distance between robot $i$ and the collision position. If the RBC is the collision between two mobile robots, then the two mobile robots will move away from the collision positions simultaneously. In this case, the distance between robot $i$ and the collision position is the half radius of robot $j$. If the RBC is the collision between robot $i$ and obstacle $j$, then only robot $i$ can move away from the collision position. In this case, the distance between robot $i$ and the collision position is the radius of obstacle $j$. $\hfill\square$
\end{remark}

\begin{figure}
    \begin{center}
	\begin{picture}(150, 50)
            \put(-45, -15){\resizebox{84mm}{21.9mm}{\includegraphics[width=2.5in]{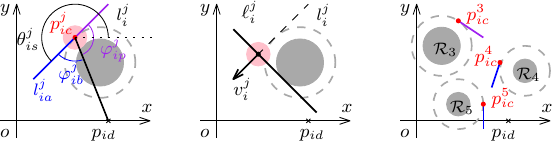}}}
	\end{picture}
    \end{center}
    \caption{Illustration of designing the imposed impulse and local controller. (\textbf{Left}) The selection of the motion direction $\theta_{is}^{j}$. (\textbf{Middle}) The exclusion of the chattering phenomenon. (\textbf{Right}) The exclusion of the deadlock phenomenon. The pink disk is robot $i$, the grey disk is $\mathcal{R}_{j}$ and the grey dashed circle is the positions where robot $i$ collides with $\mathcal{R}_{j}$.}
    \label{fig-3}
\end{figure}

%---------------------------------------------------------------------------------------------------
\subsection{Design of Imposed Impulse and Local Controller}
\label{subsec-designofimpulseinputandlocalcontroller}
%---------------------------------------------------------------------------------------------------
	
In the proposed control redesign strategy, the first two steps for the re-collision avoidance are related to each other, and are established in this subsection. The main idea is to design an impulse $\xi_{i}^{j}\in\mathbb{R}^{3}$ such that the motion direction of robot $i$ is changed to a given value $\theta_{is}^{j}\in\mathbb{R}$, and to design a local controller $u_{i}^{j}\in\mathbb{U}$ such that robot $i$ moves along $\theta_{is}^{j}$.
	
\subsubsection*{\textbf{Design of Imposed Impulse}} 
At the time instant of the RBC, the motion direction of robot $i$ is changed to the given value $\theta_{is}^{j}$ by imposing the following impulse:
\begin{align}
    \label{eqn-21}
    \xi_{i}^{j} := (0,0,\theta_{i}^{j}) = (0,0,\theta_{is}^{j}-\theta_{i}^{+}),
\end{align}
where $\theta_{i}^{+}$ is obtained from \eqref{eqn-17}. The selection of $\theta_{is}^{j}$ is given below and is also shown in the left figure of Fig.~\ref{fig-3}.
\begin{enumerate}[a)]
    \item At the collision position $p_{ic}^{j}$, the tangent line of the circle $(x-x_{j})^{2}+(y-y_{j})^{2}=(r_{i}+r_{j})^{2}$ is
    \begin{align*}
        l_{i}^{j}: (y_{j}-y_{ic}^{j})(y-y_{ic}^{j})=-(x_{j}-x_{ic}^{j})(x-x_{ic}^{j}),~x\in\mathbb{R}.
    \end{align*}
    The line $l_{i}^{j}$ can be divided into two rays by $p_{ic}^{j}$ (see the left figure of Fig. \ref{fig-3}). If $y_{j}\ne y_{ic}^{j}$, then the first ray is $y=\kappa_{i}^{j}(x-x_{ic}^{j})+y_{ic}^{j}, x\in(-\infty,x_{ic}^{j}]$ and the second ray is $y=\kappa_{i}^{j}(x-x_{ic}^{j})+y_{ic}^{j},x\in[x_{ic}^{j},+\infty)$, where $\kappa_{i}^{j}=-\frac{x_{j}-x_{ic}^{j}}{y_{j}-y_{ic}^{j}}$. If $y_{j}=y_{ic}^{j}$, then the first ray is $x=x_{ic}^{j},y\in(-\infty,y_{ic}^{j}]$ and the second ray is $x=x_{ic}^{j},y\in[y_{ic}^{j},+\infty)$.
		
    \item The collision position $p_{ic}^{j}$ and the target position $p_{id}$ are connected as the segment $p_{ic}^{j}p_{id}$. Let $\varphi_{ib}^{j},\varphi_{ip}^{j}\in[0,\pi)$ be the angles between $p_{ic}^{j}p_{id}$ and the two rays, respectively. The relation between $\varphi_{ib}^{j}$ and $\varphi_{ip}^{j}$ is $\varphi_{ib}^{j}+\varphi_{ip}^{j}=\pi$. 
		
    \item If $\varphi_{ib}^{j}\le\varphi_{ip}^{j}$, then the first ray is taken as the ray $\mathbf{l}_{i}^{j}$. Otherwise, the second ray is taken as the ray $\mathbf{l}_{i}^{j}$. Let $\varphi_{i}^{j}\in[0,\pi)$ be the angle between $p_{ic}^{j}p_{id}$ and $\mathbf{l}_{i}^{j}$. We obtain
    \begin{align}
	\label{eqn-22}
        \varphi_{i}^{j}=\min\{\varphi_{ib}^{j},\varphi_{ip}^{j}\}.
    \end{align}
    
    \item Based on the ray $\mathbf{l}_{i}^{j}$, the value of $\theta_{is}^{j}$ is set below. 
    \begin{align}
	\label{eqn-23}
	\theta_{is}^{j}=\left\{\begin{aligned}
		&\pi+\arctan(\kappa_{i}^{j}), 
            &&\text{if } y_{j}\ne y_{ic}^{j}, \varphi_{i}^{j}=\varphi_{ib}^{j}, \\
		&\arctan(\kappa_{i}^{j}),
            &&\text{if } y_{j}\ne y_{ic}^{j}, \varphi_{i}^{j}=\varphi_{ip}^{j}, \\
		&1.5\pi, 
            &&\text{if } y_{j}=y_{ic}^{j}, \varphi_{i}^{j}=\varphi_{ib}^{j}, \\
		&0.5\pi,
            &&\text{if } y_{j}=y_{ic}^{j}, \varphi_{i}^{j}=\varphi_{ip}^{j}.
	\end{aligned}\right.
    \end{align}
		
    For $i,j\in\{1,2\}$ and $i\ne j$, if $\theta_{is}^{j}=\theta_{js}^{i}$, then we choose $\theta_{is}^{j}$ as $\theta_{is}^{j}+\pi$ or choose $\theta_{js}^{i}$ as $\theta_{js}^{i}+\pi$ such that the motion directions of the two mobile robots are oriented in different directions and $\varphi_{i}^{j}+\varphi_{j}^{i}$ is minimal.
\end{enumerate}
	
From the above discussion, $\theta_{is}^{j}$ is selected as the initial motion direction after the RBC. \eqref{eqn-22} ensures that $\theta_{is}^{j}$ is selected such that the mobile robot tends to move towards the target position. In addition, since $|\theta_{is}^{j}-\theta_{js}^{i}|=\pi$ or $\mathcal{R}_{j}$ is static, along the motion direction $\theta_{is}^{j}$, robot $i$ will move away from $\mathcal{R}_{j}$. To this end, we design a local controller to guarantee robot $i$ to move along the direction $\theta_{is}^{j}$. 

\subsubsection*{\textbf{Design of Local Controller}}
After imposing the impulse $\xi_{i}^{j}$, robot $i$ moves along the motion direction $\theta_{is}^{j}$ by adopting the following local controller:
\begin{align}
    \label{eqn-24}
    u_{i}^{j}:=(v_{i}^{j},0),
\end{align}
where $u_{i}^{j}\in\mathbb{U}^{\prime}:=(0,M_{v}]\times\{0\}\subset\mathbb{U}$. Here, we do not set an exact value for $v_{i}^{j}$ but provide a constraint on $v_{i}^{j}$. In this way, $v_{i}^{j}$ can be either a constant or a time-varying function on $(0,M_{v}]$.
Let $t_{i}^{j}>0$ be the duration of using the local controller $u_{i}^{j}$. Based on \eqref{eqn-20b} and \eqref{eqn-24}, the duration $t_{i}^{j}$ can be derive from the following equation
\begin{align}
    \label{eqn-25}
    \int_{0}^{t_{i}^{j}}v_{i}^{j}dt= \left\{
    \begin{aligned}
	&0.5r_{j},&&\text{if }j\in\{1,2\}\text{ and }j\ne i, \\
	&r_{j}, &&\text{if }j\in\{3,\cdots,n\}.
    \end{aligned}\right.
\end{align}
	
Under the local controller $u_{i}^{j}$, robot $i$ moves along the motion direction $\theta_{is}^{j}$ to satisfy \eqref{eqn-20}. Once \eqref{eqn-20} is satisfied, the predefined controller $u_{i}^{\textrm{pre}}$ is re-activated to ensure the task accomplishment. At the re-activated position, the tangent line of the circle $(x-x_{j})^{2}+(y-y_{j})^{2} = r_{j}^{2}$ is the line $\ell_{i}^{j}$, which partitions the 2-D workspace into two half-plane; see the middle figure of Fig.~\ref{fig-3}. The proposed control redesign strategy ensures that robot $i$ moves towards the half-plane without $\mathcal{R}_{j}$. In this case, under the angular velocity $w_{i}^{\textrm{pre}}$ of the predefined controller $u_{i}^{\textrm{pre}}$, the motion direction will not jump but change slowly, thereby avoiding the successive collision with $\mathcal{R}_{j}$. Hence, the chattering or Zeno phenomena can be excluded.

After any RBC, robot $i$ will continue to move and may collide with other rigid bodies under the predefined controller $u_{i}^{\textrm{pre}}$. In this case, the proposed control redesign strategy is implemented repeatedly to resolve all collision occurrence. Since the selected motion direction $\theta_{is}^{j}$ in \eqref{eqn-23} is closer to the direction pointing towards the target position, robot $i$ still approaches the target position while resolving all collision occurrence, as shown in the right figure of Fig.~\ref{fig-3}. In this way, the mobile robot will not return to collide with the same rigid body again. That is, the re-collision avoidance is ensured. Further, the deadlock phenomena of repeatedly colliding with several rigid bodies can be excluded.
	
%-------------------------------------------------------------------------------------------------------
\subsection{Task Accomplishment}
\label{subsec-taskaccomplishment}
%-------------------------------------------------------------------------------------------------------
	
After resolving all collision occurrences, we show the task accomplishment under the proposed motion control strategy in this subsection. We define $z_{i}:=(\xi_{i},q_{i})\in\mathbb{Z}:=\mathbb{X}\times\{0,1\},i\in\{1,2\}$, and formulate the hybrid system below.
\begin{subequations}
    \label{eqn-26}
    \begin{align}
	\label{eqn-26a}
        \dot{z}_{i} &= \mathcal{F}_{i}(z_{i},u_{i}), &&(z_{i},u_{i}) \in\mathcal{C}_{i}^{\prime}:= \mathcal{C}_{i1}\cup\mathcal{C}_{i2},\\
	\label{eqn-26b}
        z^{+}_{i} &= \mathcal{J}_{i}(z_{i},u_{i}), &&(z_{i},u_{i}) \in\mathcal{D}_{i}^{\prime}:= \mathcal{D}_{i1}\cup\mathcal{D}_{i2},
    \end{align}
\end{subequations}
where  
$\mathcal{C}_{i1}:= \{z_{i}\in\mathbb{Z}, u_{i}\in\mathbb{U}: q_{i}=0, (\xi_{i},u_{i})\in\mathcal{C}_{i}\} \cup \{z_{i}\in\mathbb{Z}, u_{i}\in\mathbb{U}: q_{i}=0, (\xi_{i},u_{i})\in\mathcal{D}_{i}, \theta_{i}=\theta_{i}^{+}\}$, $\mathcal{C}_{i2}:=\{z_{i}\in\mathbb{Z}, u_{i}\in\mathbb{U}: q_{i}=1, \theta_{i}=\theta_{is}^{j}, u_{i}\in\mathbb{U}^{\prime}, \eqref{eqn-20} \text{ is not satisfied}\}$,
$\mathcal{D}_{i1}:= \{z_{i}\in\mathbb{Z}, u_{i}\in\mathbb{U}: q_{i}=0, (\xi_{i},u_{i})\in\mathcal{D}_{i}, \theta_{i}\ne\theta_{i}^{+}\}$, $\mathcal{D}_{i2}:= \{z_{i}\in\mathbb{Z}, u_{i}\in\mathbb{U}: q_{i}=1,  \theta_{i}=\theta_{is}^{j}, \eqref{eqn-20} \text{ is satisfied}\}$,
and $\mathcal{C}_{i}$ and $\mathcal{D}_{i}$ are defined in Section \ref{subsec-modellingofRBC}. The maps $\mathcal{F}_{i}$ and $\mathcal{J}_{i}$ are respectively defined as
\begin{subequations}
    \label{eqn-27}
    \begin{align}
	\label{eqn-27a}
	\mathcal{F}_{i}(z_{i},u_{i})&:=\left\{
	\begin{aligned}
            &\mathcal{F}_{i1}(z_{i},u_{i}),\quad &&(z_{i},u_{i})\in\mathcal{C}_{i1}, \\
            &\mathcal{F}_{i2}(z_{i},u_{i}),\quad &&(z_{i},u_{i})\in\mathcal{C}_{i2},
	\end{aligned}\right. \\
	\label{eqn-27b}
	\mathcal{J}_{i}(z_{i},u_{i})&:=\left\{
	\begin{aligned}
            &\mathcal{J}_{i1}(z_{i},u_{i}),\quad &&(z_{i},u_{i})\in\mathcal{D}_{i1}, \\
            &\mathcal{J}_{i2}(z_{i},u_{i}),\quad &&(z_{i},u_{i})\in\mathcal{D}_{i2}
	\end{aligned}\right.
    \end{align}
\end{subequations}
with $\mathcal{F}_{i1}(z_{i},u_{i}) := 
(g_{i}(\xi_{i})u_{i}^{\textrm{pre}},0), \mathcal{F}_{i2}(z_{i},u_{i}) := 
(g_{i}(\xi_{i})u_{i}^{j},$ $1), 
\mathcal{J}_{i1}(z_{i},u_{i}) := (\xi_{i}^{+}+\xi_{i}^{j},1)$ and $\mathcal{J}_{i2}(z_{i},u_{i}) := (\xi_{i},0)$. The sets $\mathcal{C}_{i}^{\prime}, \mathcal{D}_{i}^{\prime}$ are closed, and the maps $\mathcal{F}_{i}, \mathcal{J}_{i}$ are continuous. 
	
In the hybrid model \eqref{eqn-26}, the proposed motion control strategy is of a switched impulsive form with two switching modes and two impulsive modes. The switching modes come from the predefined controller \eqref{eqn-6} and the local controller \eqref{eqn-24}, while the impulsive modes come from the condition of control redesign in Section \ref{subsec-necessityandstrategyofcontrolredesign} and the conditions in \eqref{eqn-20}. Under the proposed motion control strategy, the following theorem is presented to show the task accomplishment.

\begin{theorem}
    \label{thm-2}
    Under allowable collisions, consider robot $i$ with the hybrid model \eqref{eqn-26}. Given any initial state $z_{i0}\in\mathbb{X}\times\{0\}$, the state $z_{i}$ eventually converges to $(\xi_{id},0)$, where $\xi_{id}\in\mathbb{X}$ is the target state of robot $i$.
\end{theorem}

\begin{IEEEproof}
	Under allowable collision, if there do not exist RBCs or if there exist the RBCs without changing the motion direction, then the state of robot $i$ evolves by the map $\mathcal{F}_{i1}$ with the predefined controller $u_{i}^{\textrm{pre}}$ in \eqref{eqn-6}, and thus the task accomplishment can be guaranteed, which is the trivial case. Next, we consider the non-trivial case, where the RBC occurs and the motion direction is changed. In this case, the proposed control redesign strategy is used. That is, the impulse $\xi_{i}^{j}$ in \eqref{eqn-23} is imposed and the local controller $u_{i}^{j}$ in \eqref{eqn-24} is activated to drive robot $i$ away from the collision position, which corresponds to the state evolution by the maps $\mathcal{J}_{i1}$ and $\mathcal{F}_{i2}$. Once the conditions in \eqref{eqn-20} are satisfied, the predefined controller $u_{i}^{\textrm{pre}}$ is re-activated, which corresponds to the state evolution by the maps $\mathcal{J}_{i2}$ and $\mathcal{F}_{i1}$. Since the number of the obstacles is finite and the re-collision with the same rigid body is avoided by the proposed control redesign strategy, the number of the collisions is also finite. From \eqref{eqn-25}, we see that the local controller $u_{i}^{j}$ is activated for a short time. Hence, the finite collisions can be resolved within finite time. That is, the state evolves by the maps $\mathcal{J}_{i1},\mathcal{F}_{i2}$ and $\mathcal{J}_{i2}$ within finite time, and thus the mode of the hybrid model \eqref{eqn-26} will be switched to the mode with the map $\mathcal{F}_{i1}$ eventually. In this way, for any initial state $z_{i0}\in\mathbb{X}\times\{0\}$, the state convergence to $(\xi_{id},0)$ will be guaranteed by the map $\mathcal{F}_{i1}$.
\end{IEEEproof} 

From Theorem 2, using the proposed motion control strategy, each mobile robot will converge to its target state. Hence, the tasks of the two mobile robots under allowable collisions are accomplished.

\begin{figure}[!t]
	\subfigure[The predefined controller]{
		\begin{minipage}[t]{0.47\linewidth}
			\begin{center}
				\begin{picture}(70,90)
				\put(-30,-4){\resizebox{44.4mm}{33.3mm}{\includegraphics[width=2.5in]{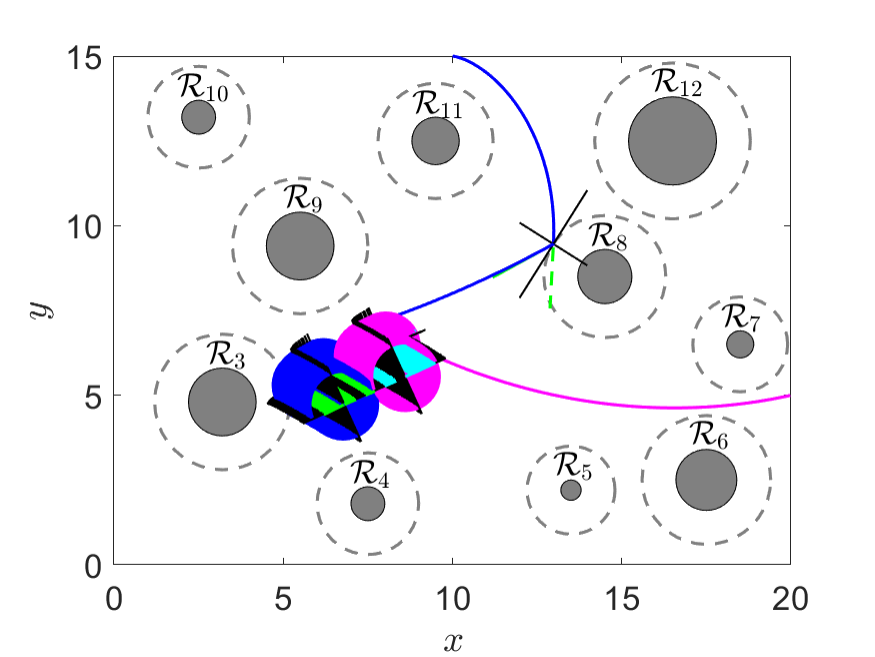}}}
				\end{picture}
			\end{center}
	\end{minipage}}
	\subfigure[The proposed control strategy]{
		\begin{minipage}[t]{0.47\linewidth}
			\begin{center}
				\begin{picture}(70,90)
				\put(-30,-4){\resizebox{44.4mm}{33.3mm}{\includegraphics[width=2.5in]{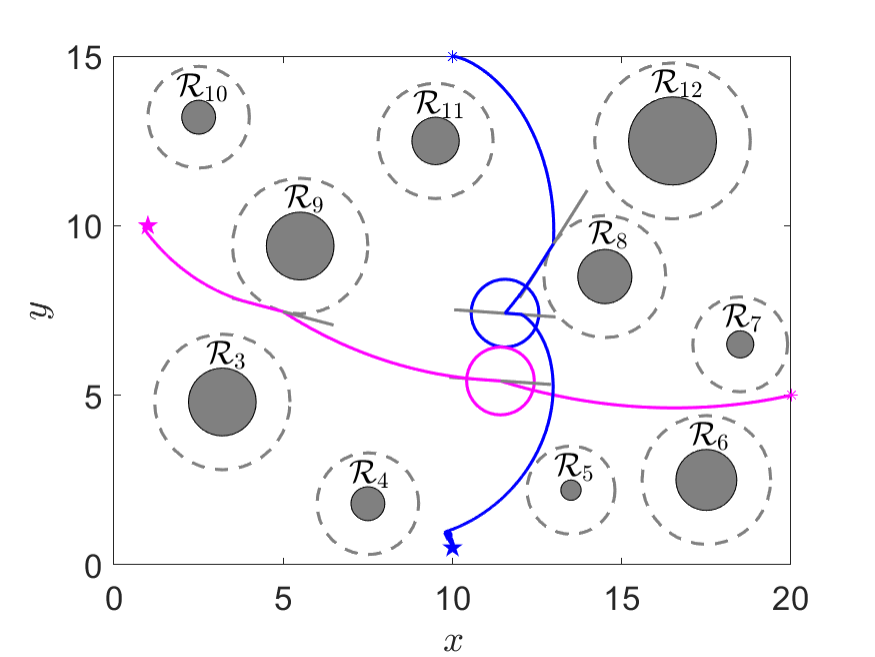}}}
				\end{picture}
			\end{center}
	\end{minipage}}
	\vspace{-0.25cm}
	\caption{Illustration of the position trajectories of the two mobile robots in the workspace. The blue and magenta indicate robot 1 and robot 2, respectively. In (a), the black solid segments are the coordinate axes of the $lcf$. In (b), the grey segments are the tangent lines at collision positions.}
	\label{fig-4}
\end{figure}

%%%%%%%%%%%%%%%%%%%%%%%%%%%%%%%%%%%%%%%%%%%%%%%%%%%%%%%%%%%%%%%%%%%%%%%%%%%%%%%%%%%%%%%%%%%%%%%%%%%%%%%%
\section{Numerical Results}
\label{sec-simulation}
%%%%%%%%%%%%%%%%%%%%%%%%%%%%%%%%%%%%%%%%%%%%%%%%%%%%%%%%%%%%%%%%%%%%%%%%%%%%%%%%%%%%%%%%%%%%%%%%%%%%%%%%

In this section, a numerical example is presented to illustrate the derived control strategy. All simulations are carried out via MATLAB R2022b on a laptop with AMD R7-5800H and 16GB RAM. We consider a workspace $\mathbb{S}=[0,20]\times[0,15]$ with 10 static obstacles (see the gray regions in Fig. \ref{fig-4}). For two mobile robots with the dynamics \eqref{eqn-5}, the radii are $1$, the initial states are respectively $\xi_{10}=(10, 15,\pi)$ and $\xi_{20}=(20, 5, 0.07\pi)$, and the target states are respectively $\xi_{1d}=(10, 0.5, 0.1\pi)$ and $\xi_{2d}=(1, 10, -0.3\pi)$. 

To accomplish the tasks, the predefined controller \eqref{eqn-6} is applied and the parameters are set as $\rho=9, \sigma_{1}=1.25, \sigma_{2}=0.6, \sigma_{3}=1.2, M_{v}=10$ and $M_{w}=2$. Note that the collision avoidance is embedded in the design of the predefined controller \eqref{eqn-6}. However, the tasks are still not accomplished if the predefined controller \eqref{eqn-6} is implemented only. In this case, the position trajectories of the two mobile robots are depicted in Fig.~\ref{fig-4}(a). We can see from Fig.~\ref{fig-4}(a) that robot 1 collides with the obstacle $\mathcal{R}_{8}$ first, and then collide with robot 2 successively, thereby resulting in the chattering phenomena. In particular, after the 762th collision, all collisions between the two mobile robots occur at the same position, which implies the deadlock phenomenon. Hence, we conclude that the tasks cannot be accomplished under the predefined controller \eqref{eqn-6} only, and it is necessary to redesign the control strategy. 

\begin{figure}[!t]
	\begin{center}
		\begin{picture}(110, 130)
		\put(-85, -15){\resizebox{98mm}{49mm}{\includegraphics[width=2.5in]{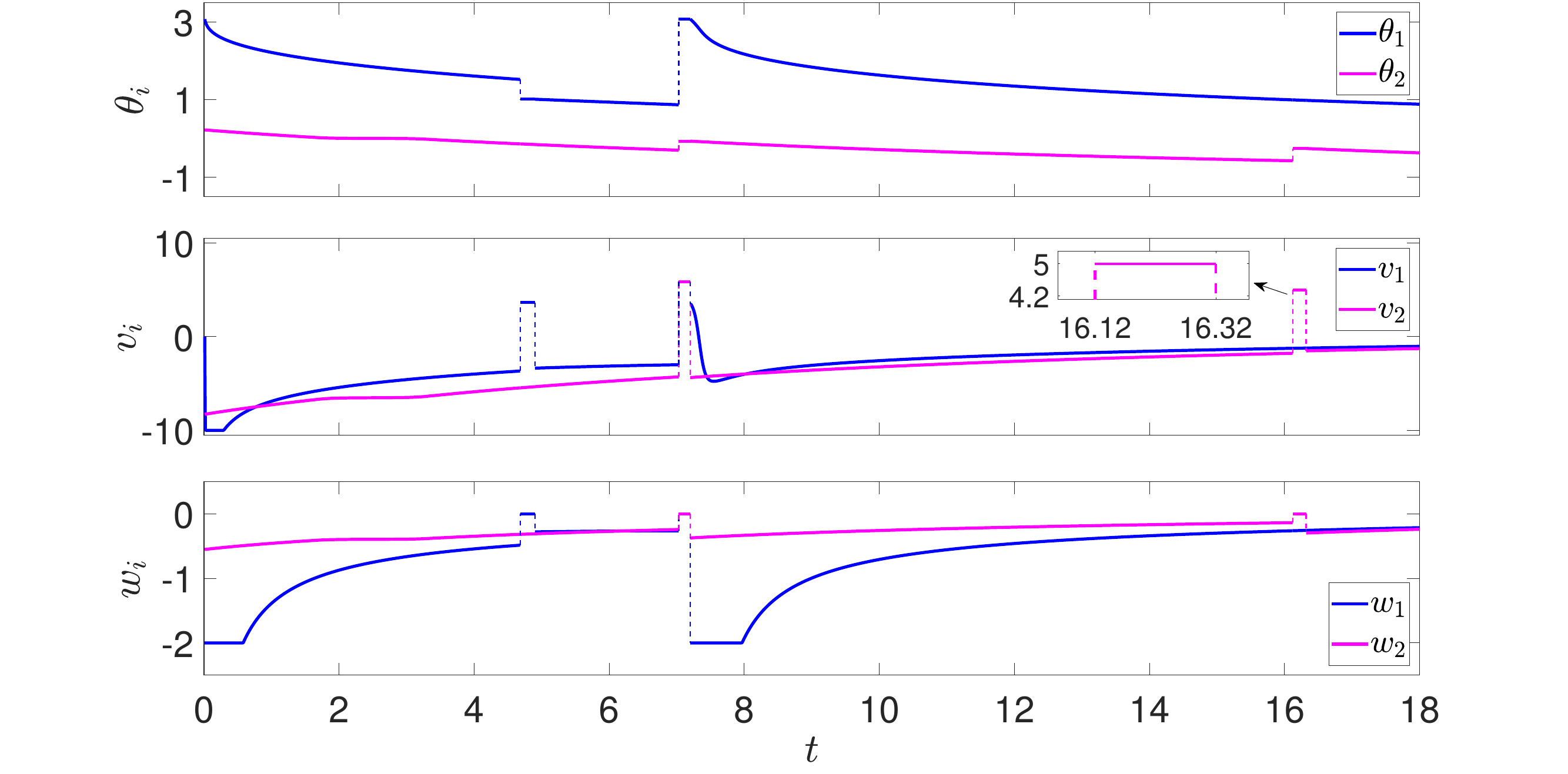}}}
		\end{picture}
	\end{center}
	\caption{Illustration of the motion directions and control inputs of the two mobile robots with the proposed control strategy in $t\in[0,18]$.}
	\label{fig-5}
\end{figure}

We next illustrate the proposed control strategy, and apply the three-step control redesign strategy in Section \ref{subsec-necessityandstrategyofcontrolredesign} after each RBC. Here we set the linear velocities in the local controllers \eqref{eqn-24} as constants. The position trajectories of the two mobile robots are shown in Fig.~\ref{fig-4}(b), where the target positions are reached after two RBCs. That is, the tasks of the two mobile robots are accomplished under the proposed control strategy. Comparing Fig.~\ref{fig-4}(a) and Fig.~\ref{fig-4}(b), we can see the advantages of the proposed control strategy: (i) the task accomplishment is guaranteed; (ii) the re-collision with the same rigid body is avoided; (iii) the chattering and deadlock phenomena are excluded.

To show the effects of the imposed impulses and the local controllers, the motion directions and control inputs in $[0,18]$ are presented in Fig.~\ref{fig-5}. We can see that for each mobile robot, an RBC causes a jump of the motion direction and two jumps of the control input. The jump of the motion direction and the first jump of the control input occur at the collision time. 
Then the local controller is imposed for a short time and the second jump of the control input occurs. For example, robot 2 collides with the obstacle $\mathcal{R}_{9}$ at $t=16.12$s. We set the linear velocity $v_{2}^{9}=5$ and derive the distance $r_{9}=1$ from \eqref{eqn-20b}. From \eqref{eqn-25}, we obtain the duration $t_{2}^{9}=0.2$s. Hence, $\theta_{2}$ jumps at $t=16.12$s, while $(v_{2},w_{2})$ jumps at $t=16.12$s and $t=16.32$s, as shown in Fig.~\ref{fig-5}. Since the velocity in the local controller \eqref{eqn-24} is constant, the motion direction and control input do not change when the local controller is implemented; see the time interval $[16.12, 16.32]$ in Fig.~\ref{fig-5}. Similarly, we can compute all the jump times and observe the constant local controller \eqref{eqn-24}.
	
%%%%%%%%%%%%%%%%%%%%%%%%%%%%%%%%%%%%%%%%%%%%%%%%%%%%%%%%%%%%%%%%%%%%%%%%%%%%%%%%%%%%%%%%%%%%%%%%%%%%%%%%
\section{Conclusion}
\label{sec-conclusion}
%%%%%%%%%%%%%%%%%%%%%%%%%%%%%%%%%%%%%%%%%%%%%%%%%%%%%%%%%%%%%%%%%%%%%%%%%%%%%%%%%%%%%%%%%%%%%%%%%%%%%%%%%

In this letter, the motion control problem was studied for two mobile robots under allowable collisions. To solve this problem, for the first time, the dynamics of each mobile robot under allowable collisions was formulated into a hybrid system. We discussed the necessity of the controller redesign, and applied the impulsive control techniques to propose a control redesign strategy. The proposed control strategy guarantees the task accomplishment and the re-collision avoidance with the same rigid body, thereby excluding the chattering and deadlock phenomena. Future work will focus on the motion control design of multiple mobile robots under allowable collisions.

\end{document}